\documentclass[10pt,twocolumn,letterpaper]{article}

\usepackage{wacv}              

\usepackage{graphicx}
\usepackage{amsmath}
\usepackage{amssymb}
\usepackage{booktabs}
\pdfmapfile{+times.map}

\usepackage[pagebackref,breaklinks,colorlinks]{hyperref}

\usepackage[capitalize]{cleveref}
\crefname{section}{Sec.}{Secs.}
\Crefname{section}{Section}{Sections}
\Crefname{table}{Table}{Tables}
\crefname{table}{Tab.}{Tabs.}

\def\wacvPaperID{1826} 
\def\confName{WACV}
\def\confYear{2025}

\usepackage{xspace}
\def\methodNAME{TROY-VIS\xspace} 
\usepackage{booktabs}
\usepackage{graphicx}
\usepackage{paralist}
\usepackage{multirow}
\usepackage{pifont}
\newcommand{\cmark}{\ding{51}}%

\begin{document}

\title{Towards Real-Time Open-Vocabulary Video Instance Segmentation}

\author{Bin Yan$^{1}$\thanks{This work was performed while Bin Yan worked as an intern at
Google.},
Martin Sundermeyer$^{2}$,
David Joseph Tan$^{2}$, 
Huchuan Lu$^{1}$,
Federico Tombari$^{2,3}$\\
$^{1}$ Dalian University of
Technology
$^{2}$ Google
$^{3}$ TU Munich
}

\maketitle

\begin{abstract}

In this paper, we address the challenge of performing open-vocabulary video instance segmentation (OV-VIS) in real-time. We analyze the computational bottlenecks of state-of-the-art foundation models that performs OV-VIS, and propose a new method, \methodNAME, that significantly improves processing speed while maintaining high accuracy. We introduce three key techniques: (1) Decoupled Attention Feature Enhancer to speed up information interaction between different modalities and scales; (2) Flash Embedding Memory for obtaining fast text embeddings of object categories; and, (3) Kernel Interpolation for exploiting the temporal continuity in videos. Our experiments demonstrate that \methodNAME achieves the best trade-off between accuracy and speed on two large-scale OV-VIS benchmarks, BURST and LV-VIS, running $20\times$ faster than GLEE-Lite (25 FPS v.s. 1.25 FPS) with comparable or even better accuracy. These results demonstrate \methodNAME's potential for real-time applications in dynamic environments such as mobile robotics and augmented reality. Code and model will be released at \url{https://github.com/google-research/troyvis}.

\end{abstract}

\section{Introduction}

In recent years, open-vocabulary video instance segmentation (OV-VIS) has gained increasing attention in mobile robotics, autonomous driving, and augmented reality where diverse objects need to be tracked in dynamic environments. Different from traditional video instance segmentation~\cite{MaskTrackRCNN,ytvis21dataset,ovis} which only focuses on a limited set of pre-defined object categories, OV-VIS~\cite{burst,LVVIS} requires the methods to identify, segment, and track objects of unlimited categories outside the supervised training set. This presents significant challenges for traditional models in computer vision. 
However, the rapid development of foundation models~\cite{CLIP,SAM,SAM2,GLEE} in recent years has brought new opportunities. Trained on vast amounts of data, these models exhibit unprecedented zero-shot understanding capabilities. For example, the recent object-level foundation model GLEE~\cite{GLEE} has achieved state-of-the-art performance on several large-scale OV-VIS benchmarks, including LV-VIS~\cite{LVVIS} and BURST~\cite{burst}.

\begin{figure}[tb]
\centering
\includegraphics[width=\linewidth]{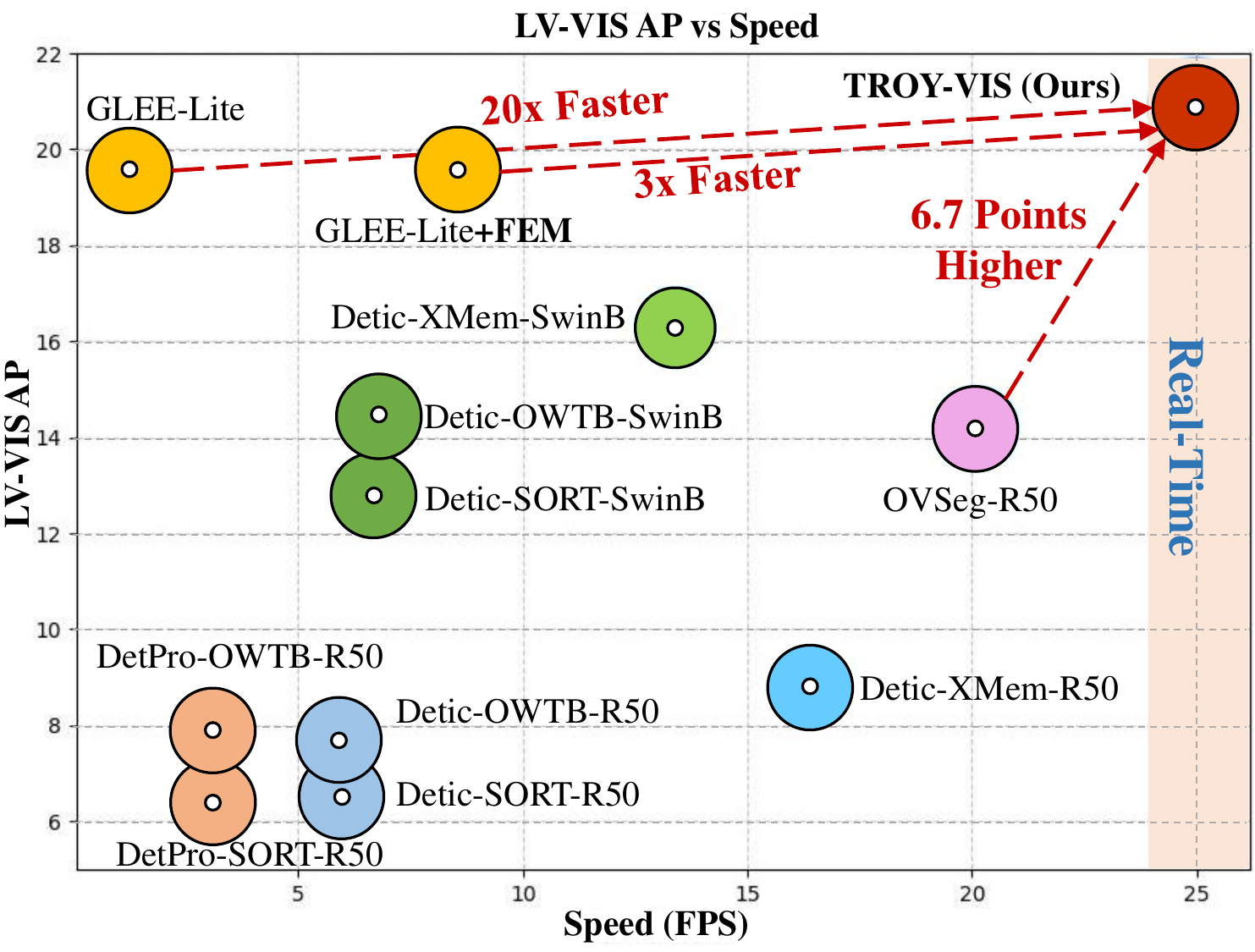}
\caption{Performance and speed comparison on the LV-VIS~\cite{LVVIS} benchmark. \methodNAME is the only method that runs in real-time. Compared with GLEE-Lite~\cite{GLEE} \methodNAME runs $20\times$ faster while achieving better results. \methodNAME surpasses OVSeg-R50~\cite{LVVIS}, the previously fastest model, by 6.7 AP. }
\label{fig:teaser}
\end{figure}

Despite the progress made, existing approaches for OV-VIS still face significant limitations. The most notable issue is the computational inefficiency of state-of-the-art models like GLEE~\cite{GLEE}, which achieve impressive accuracy but are unable to process sensor streams in real-time. GLEE-Lite, the most lightweight variant of GLEE, processes video frames at only 1.25 frames per second (FPS) on an A100 GPU, far below the threshold (24 FPS) required for real-time applications as shown in Fig.~\ref{fig:teaser}. This poor efficiency restricts the deployment of these models in time-sensitive real-world scenarios, where both accuracy and speed are equally crucial. Consequently, to make OV-VIS methods applicable to real-world scenarios, the greatest challenge is reducing their substantial computational burden.

In this paper, we propose a novel architecture and inference strategy \textbf{T}owards \textbf{R}eal-time \textbf{O}pen-vocabular\textbf{Y} \textbf{V}ideo \textbf{I}nstance \textbf{S}egmentation (\methodNAME). First, we analyze every component of GLEE, identifying the text encoder, feature enhancer, and instance decoder as the most important computational bottlenecks. Based on this analysis, we then introduce three core techniques to improve processing efficiency: (1) Decoupled Attention Feature Enhancer, which replaces heavy modality-scale hybrid attention with efficient modality attention and scale attention, reducing computational and memory costs; (2) Flash Embedding Memory, which allows for fast retrieval of text embeddings without redundant recalculations during training and inference; and, (3) Kernel Interpolation, which leverages the temporal continuity in videos to obtain proxy instance kernels by interpolating accurate kernels from key frames. These innovations allow \methodNAME to achieve significant speed improvements up to $20\times$ without sacrificing accuracy.

Our experimental results demonstrate that \methodNAME outperforms other efficient methods in both speed and accuracy on large-scale OV-VIS benchmarks, BURST~\cite{burst} and LV-VIS~\cite{LVVIS}. Specifically, \methodNAME runs at 25 FPS ($20\times$ faster than GLEE-Lite), while achieving state-of-the-art performance on these datasets. To the best of our knowledge, we propose the first real-time open-vocabulary video instance segmentation model, paving the way for the application of OV-VIS to real-world scenarios.

\section{Related Works}
\subsection{Open-Vocabulary Perception}
Compared with traditional closed-set perception operating within a predefined and unchanging set of categories, open-vocabulary perception~\cite{OVD,LVVIS,burst} requires models to have strong generalization ability so as to recognize categories unseen during training. Since this ability allows models to operate more effectively in complex, dynamic, and unpredictable environments, open-vocabulary perception has drawn more and more attention in recent years.

In this field, the primary focus has been on image-level open-vocabulary detection (OVD)~\cite{OVD}. Inspired by the great success of large-scale image-text pretraining~\cite{CLIP}, recent studies redefine open-vocabulary object detection as a region-text matching task and utilize large-scale image-text datasets to expand the vocabulary. GLIP~\cite{GLIP} first unifies object detection and phrase grounding, introducing a grounded pre-training framework that shows superior capabilities in a zero-shot setting. OWL-ViT~\cite{OWL-ViT} extends vision transformers to simple, yet effective, open-vocabulary detectors by fine-tuning them with detection and grounding datasets. Later works like Grounding DINO~\cite{groundingdino-1.5} and YOLO-World~\cite{yoloworld} replace the original ATSS~\cite{ATSS} architecture in GLIP~\cite{GLIP} with the DINO~\cite{dino} and the YOLO~\cite{YOLOv8} framework respectively.

Compared with OVD, open-vocabulary video instance segmentation~\cite{LVVIS,burst} (OV-VIS) raises additional challenges because it requires models to detect, segment, and track object simultaneously in a zero-shot setting. BURST~\cite{burst} and LV-VIS~\cite{LVVIS} are two challenging OV-VIS benchmarks, which cover 482 and 1196 object categories, respectively. Mainstream methods~\cite{burst,LVVIS} combine existing open-vocabulary detectors~\cite{detic} with off-the-shelf association methods~\cite{bewley2016simple,xmem} to accomplish OV-VIS. However, the separation of detection and association models may lead to suboptimal performance. Recently, GLEE~\cite{GLEE} achieves state-of-the-art performance on both OV-VIS benchmarks by training a powerful Transformer-based~\cite{maskdino} detector and associating objects based on object queries.

\subsection{Generalist Models for Vision Perception}
We recently witnessed a paradigm shift from specialist to generalist models~\cite{unicorn,tarvis,UNINEXT,xdecoder,OpenSeeD,GLEE,APE}, which can solve multiple tasks using a unified model and condense knowledge from various domains into one suite of parameters. In object tracking, Unicorn~\cite{unicorn} and TarVIS~\cite{tarvis} are the pioneers in this field. On one hand, Unicorn~\cite{unicorn} unifies the temporal correspondence required by different tracking tasks and detects objects of various properties with the help of target priors. On the other, TarVIS~\cite{tarvis} unifies multiple pixel-level tracking tasks into a query-based temporal segmentation architecture and uses different target queries to accomplish different tracking tasks.

Later, several multi-modal perception generalist models~\cite{UNINEXT,xdecoder,OpenSeeD,APE,GLEE} were developed by incorporating advanced text encoders~\cite{bert,CLIP}. With the powerful vision-language understanding abilities, they solved more complicated perception tasks. To be more specific, X-Decoder~\cite{xdecoder} is a unified framework for image-level referring and open-vocabulary segmentation. OpenSeeD~\cite{OpenSeeD} proposes a simple unified open-vocabulary detection and segmentation framework. APE~\cite{APE} is a universal image perception model, which can perform detection, segmentation, and grounding with a single model. Although achieving success in image-level perception tasks, the effectiveness of these methods on videos is unexplored. In contrast, UNINEXT~\cite{UNINEXT} reformulates 10 instance perception tasks on both images and videos into a prompt-guided object discovery and retrieval paradigm. Building on UNINEXT, GLEE~\cite{GLEE} further scales-up the training data and enhances its open-vocabulary ability, which has shown superior performance on open-vocabulary video instance segmentation~\cite{burst,LVVIS}.

\subsection{Efficient Vision Models}
In recent years, as model sizes have consistently grown~\cite{scaling-law}, there has been a qualitative leap in model performance~\cite{eva02,gpt-4,SORA}, with the emergence of previously unforeseen capabilities~\cite{CLIP,SAM}. However, the surge in the number of model parameters and computational requirements has also hindered the practical application of these models, particularly in deploying them on resource-constrained edge platforms. To address this challenge, many recent works have attempted to train efficient vision models that significantly reduce computational costs while maintaining model performance as much as possible.

Based on powerful vision-language or pure vision foundation models like CLIP~\cite{CLIP} and SAM~\cite{SAM}, there are a series of works attempting to build their efficient variants. For example, TinyCLIP~\cite{tinyclip} proposes two techniques, affinity mimicking and weight inheritance, largely reducing the model size. MobileCLIP~\cite{mobileclip} further designs more lightweight image and text encoder, optimizing them with multi-modal reinforced training. Besides, most efficient variants of SAM~\cite{SAM} try transferring knowledge from the original heavy ViT-H backbone to more lightweight ones. Specifically, MobileSAM~\cite{MobileSAM} and EfficientViT-SAM~\cite{efficientvit-sam} adopt ViT-Tiny and EfficientViT~\cite{EfficientViT} as the backbone respectively, training them via knowledge distillation using MSE Loss. In contrast, EfficientSAM~\cite{EfficientSAM} trains the lightweight encoder with more complex masked image pretraining~\cite{MAE}.
Although achieving great success, these techniques are specifically designed for architectures like CLIP~\cite{CLIP} and SAM~\cite{SAM}, being difficult to apply on other frameworks like DETR~\cite{detr}-style models.

In the field of object-level perception, DETR-style methods~\cite{dino,RT-DETR,UNINEXT,GLEE} are currently the most popular solutions. There are also some works trying to design more lightweight models of this style. For example, \cite{MobileInst,omnidet-turbo,groundingdino-1.5} design more lightweight feature enhancers and instance decoders to speed-up models. Moreover, MobileInst~\cite{MobileInst} proposes a technique called kernel reuse to save computations of the instance decoder. Our kernel interpolation uses similar principles but has two key differences. First, while MobileInst only applies kernel reuse on closed-vocabulary VIS benchmarks~\cite{MaskTrackRCNN,ytvis21dataset} with tens of object categories, we introduce kernel interpolation that can be scaled to the more challenging open-vocabulary VIS task~\cite{burst,LVVIS}. Furthermore, the kernel reuse in MobileInst requires temporal training, restricting the training data to annotated videos. In contrast, we find that solely applying kernel interpolation during the inference can already achieve good results, allowing us to train on larger scale image datasets. To optimize the text encoder, we introduce Flash Embedding Memory, which adopts a similar motivation as the language cache in the object detection method OmniDet-Turbo~\cite{omnidet-turbo}. This technique enables us to use larger text encoders without causing extra computational costs.


\section{Methodology}

\begin{figure*}[tb]
\centering
\includegraphics[width=1.0 \linewidth]{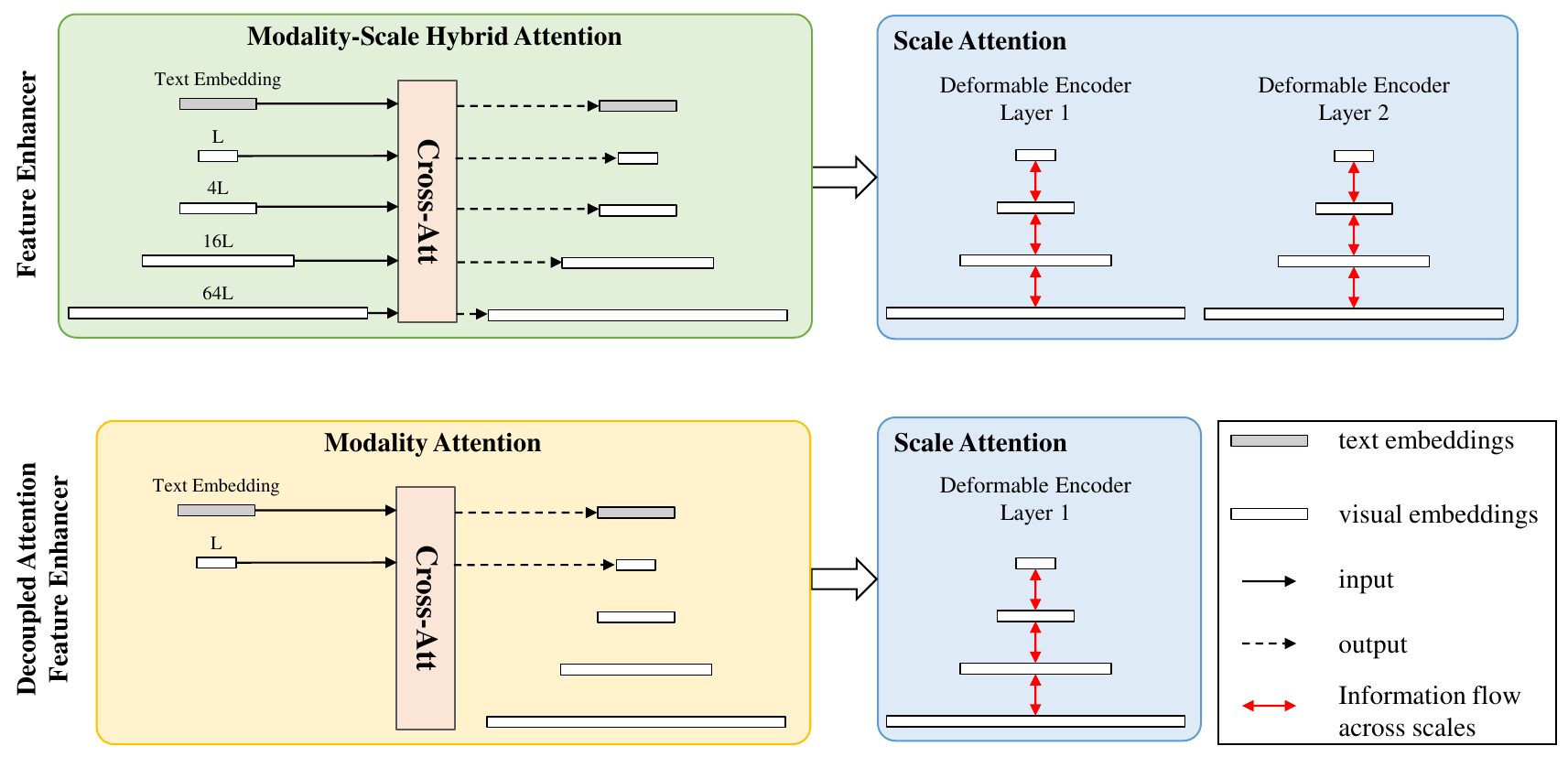}
\caption{Architecture comparison between the original feature enhancer and our decoupled attention feature enhancer. In our design, a fast modality attention and an efficient scale attention are used to replace the heavy modality-scale hybrid attention.}
\label{fig:feature_enhancer}
\end{figure*}

Existing OV-VIS methods~\cite{LVVIS,GLEE} have achieved state-of-the-art accuracy. However, their inference time is far from real-time, consequently limiting their real world application. 
%
Taking the recent object-level foundation model GLEE~\cite{GLEE} as an example, it took GLEE-Lite 805ms to process a single frame on the LV-VIS~\cite{LVVIS} dataset which is equivalent to 1.25FPS only (see Tab.~\ref{tab:efficiency}).
Note that this is the fastest version of GLEE running on a powerful A100 GPU. 
To overcome this hurdle and improve the speed, we analyzed component-wise latency in Tab.~\ref{tab:efficiency} and discovered that the text encoder, feature enhancer, and instance decoder are the most time-consuming bottlenecks. 

\paragraph{Text Encoder.} 
The initial step in OV-VIS includes having the desired object categories pass
through the text encoder to obtain the text embeddings. 
As the number of categories increases, the latency of the text encoder grows drastically.
For instance, to process 1196 categories of LV-VIS~\cite{LVVIS}, it takes the text encoder more than 800ms as shown in Tab.~\ref{tab:efficiency}.
What's worse is that the original implementation of GLEE repeats this computation on each frame, which causes serious computational redundancy.

\paragraph{Feature Enhancer.} To correlate the visual and textual features, an early fusion module, which is a cross-attention between two modalities, is introduced. Specifically, the input visual features are hierarchical representations with strides of \{8, 16, 32, 64\}. Furthermore, when processing a video frame of 480p, the total number of visual tokens reaches 8K. This amount of visual tokens leads to heavy computational costs and high memory demands during training. 

\paragraph{Instance Decoder.}\label{sec:preliminaries} The instance decoder of the previous works follows the design of Mask DINO~\cite{maskdino} but replaces the fixed-size classifier with vision-language alignment~\cite{detclip}. As for mask prediction, first, $N$ object queries $Q$ are passed through the Transformer decoder, obtaining $N$ instance kernels $K$. Then, the $N$ instance masks $m$ are predicted by convolving $N$ instance kernels $K$ with a 1/4 downsampled pixel embedding map $M$, written as
\begin{align}
    m=K \star M; ~ K=\text{Dec}(Q) ~.
    \label{eq:0}
\end{align}
The instance decoder consists of 9 decoder layers and processes $N=300$ queries. This setting makes the instance decoder computationally expensive to run on each frame.

\paragraph{\methodNAME.}
To address the aforementioned bottlenecks, we introduce three key techniques: (1)~\textbf{Decoupled Attention Feature Enhancer} to speed up information interaction between different modalities and scales; (2)~\textbf{Flash Embedding Memory} for fast retrieval of the text embeddings of object categories; and, (3)~\textbf{Kernel Interpolation} for exploiting the temporal continuity in videos and making predictions efficiently. 

\subsection{Decoupled Attention Feature Enhancer}
To alleviate the large memory and computational costs in the feature enhancer, we propose a lightweight decoupled attention feature enhancer. Our method is inspired by depthwise separable convolution~\cite{xception}, which decomposes a dense 3x3 convolution into a depth-wise 3x3 convolution in spatial dimension and a point-wise 1x1 convolution in channel dimension. Similarly, we decouple the original modality-scale hybrid attention into a modality attention and a scale attention as shown in Fig.~\ref{fig:feature_enhancer}. In this section, we introduce this design in more detail and compare it with the original feature enhancer in terms of time and space complexity.

Before computing the specific complexity of the two architectures, we first derive the general formula for the time and space complexity of the cross-attention operation. Suppose there are two input sequences with length of $L_1$ and $L_2$, both with a feature dimension of $d$. Cross-attention consists of three steps: (1) projection of query, key, and value; (2) attention matrix computation between query and key; and, (3) weighted sum on value using the attention matrix. The time complexity $Com_{T}$ and space complexity $Com_{S}$ are denoted as
\begin{align}
    Com_{T} &= 2 \times L_1 \times L_2 \times d + (L_1 + L_2) \times d^2,\\
    Com_{S} &= L_1 \times L_2 + (L_1 + L_2) \times d ~.
    \label{eq:1}
\end{align}
Usually, $L_1 >> d$ and $L_2 >> d$ especially when dealing with high-resolution images and a large vocabulary set. Thus, both $Com_T$ and $Com_S$ are nearly proportional to $L_1 \times L_2$.

In our case, the length of the text sequence is $L_t$ and there are $L_v$ tokens in the visual features at the smallest scale (stride of 64). Then the total number of visual tokens from all scales is $(1+4+16+64)L_v=85L_v$. Thus, the time and space complexity of modality-scale hybrid attention are about $170 L_t \times L_v \times d$ and $85 L_t \times L_v$, respectively. In contrast, in our proposed modality attention, only the visual features with the lowest resolution are used to compute the cross-attention with the text tokens. Thus, its time and space complexities are $2L_t \times L_v \times d$ and $ L_t \times L_v$, respectively, being $85\times$ more efficient than modality-scale hybrid attention.

However, the modality attention itself is not equivalent to the original hybrid attention since it lacks scale-level information interaction. To compensate for this drawback, we combine our modality attention with a scale attention module. Specifically, we find that the pixel decoder (\ie deformable encoder~\cite{deformableDETR}) can naturally serve as the scale attention module because it enhances the visual features by modelling the relationship among points from different scales. Therefore, we reuse the pixel encoder as the scale attention module instead of introducing additional networks. To make the scale attention more efficient, we reduce the number of deformable encoder layers from 6 to 3.

\begin{figure}[tb]
\centering
\includegraphics[width=1.0 \linewidth]{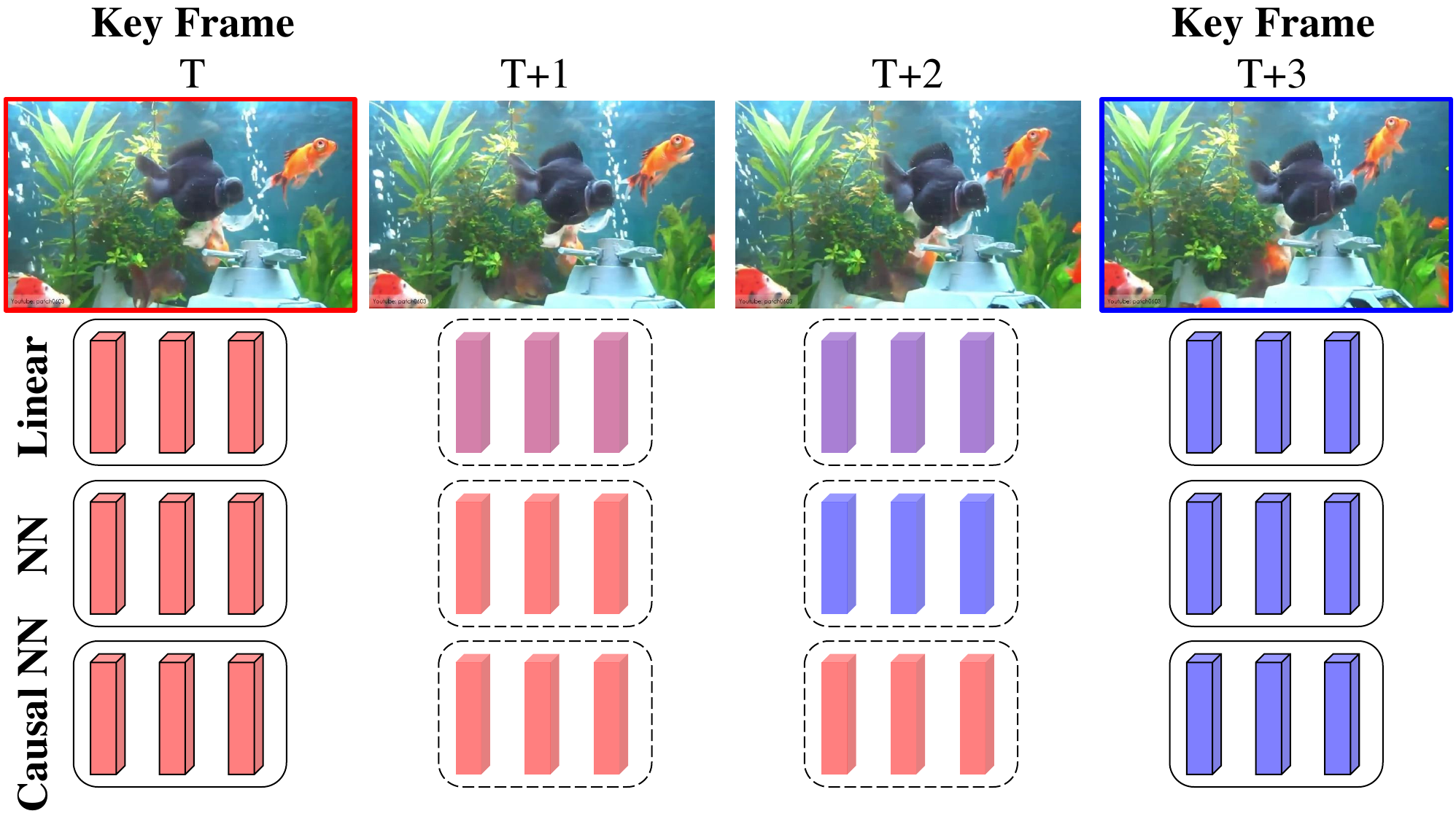}
\caption{Illustration of kernel interpolation. Cuboids represent instance kernels and kernels from the same frame are in the same color. Besides, accurate kernels on key frames and proxy kernels from non-key frames are circled in solid and dashed lines respectively. We explore three types of interpolation methods: linear, nearest neighbor (NN), and causal NN interpolation.}
\label{fig:kernel}
\end{figure}

\subsection{Flash Embedding Memory}
When dealing with a large number of vocabularies, the processing time of the language encoder can become the bottleneck of the entire network. To address this, we propose Flash Embedding Memory. Instead of repeatedly running the text encoder to get the text embeddings of object categories in the current batch on-the-fly, we store the embeddings of all seen categories in a memory, with category name as the key and text embedding as the value. In this way, each category is only processed once without repeated computations. For seen categories, the embeddings can be obtained in a flash by retrieving the keys and picking out the corresponding values in $O(1)$ time complexity.

Notably, Flash Embedding Memory can be used to speed up both inference and training (if the text encoder is frozen during training). Before training, we can first summarize all object categories from diverse training data and store them into a set without duplication. Then, we send them into the text encoder in parallel, saving the category-embedding pairs in the Flash Embedding Memory. With this powerful memory, during training, there is no need to run even to keep the text encoder because all the required text embeddings can be retrieved in a flash. 

During inference, the following two scenarios are considered: evaluating on benchmarks (\ie, categories are fixed); and, testing in the wild (\ie, categories are not fixed and can be infinite). In the former, we can construct Flash Embedding Memory of all categories in advance, then directly use the embeddings during the inference. Furthermore, in the latter case, since the possible categories are infinite, it is not possible to cover all of them in the memory beforehand. However, Flash Embedding Memory strategy can still work with some small modifications. When coming across unseen categories, we can choose $K$ nearest neighbours of the new categories from the existing ones in the memory based on their semantic similarities. The embeddings of new category can then be obtained by averaging the embeddings of its $K$ nearest neighbours. Finally, the new key-value pair is saved to the memory.

Moreover, Flash Embedding Memory allows us to use embeddings from more powerful text encoders  during inference without incurring additional computational costs. In this work, we replace the original CLIP-B with more advanced EVA-02-CLIP-L~\cite{eva02}.

\begin{figure*}[tb]
\centering
\includegraphics[width=1.0 \linewidth]{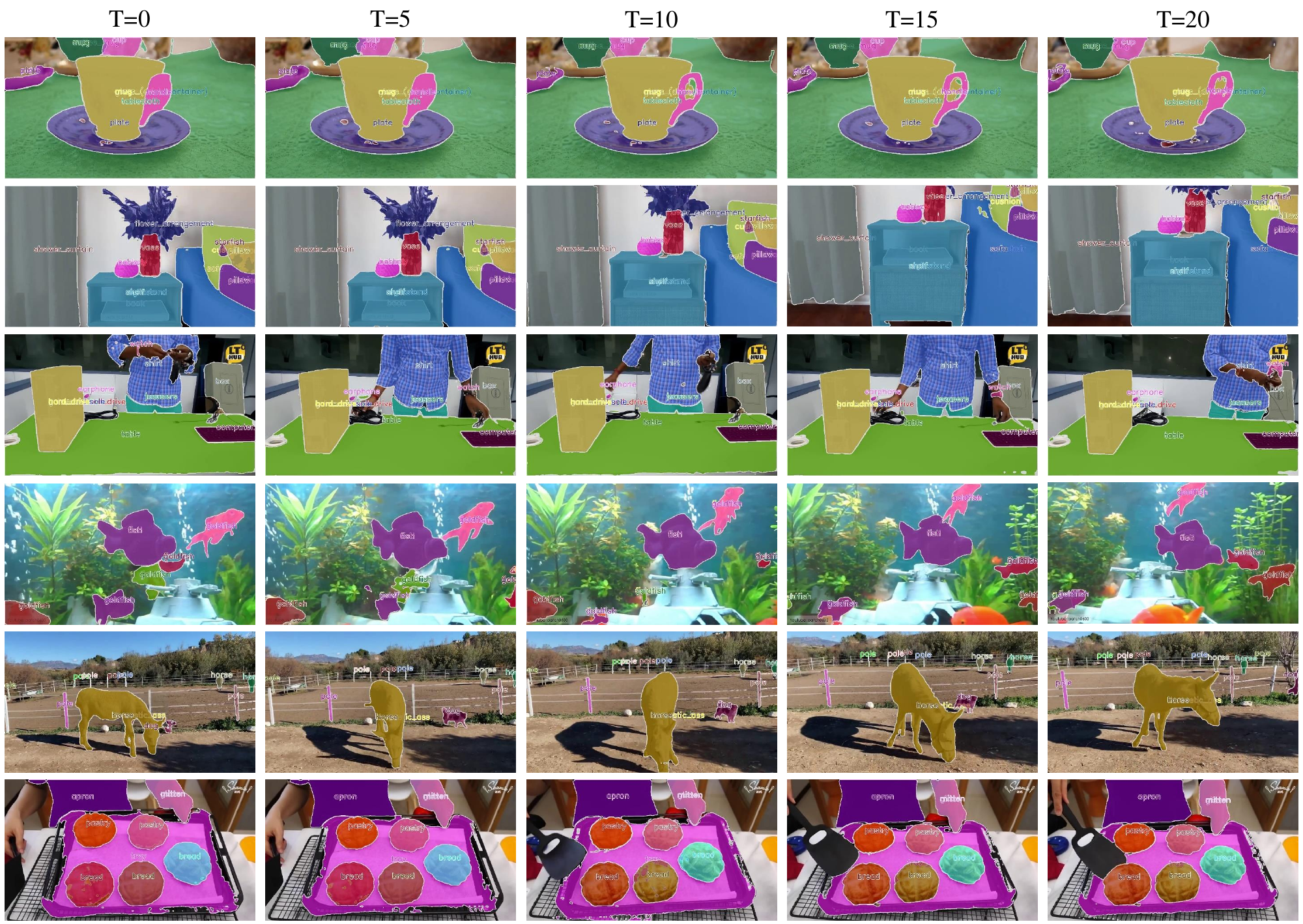}
\caption{Qualitative results of \methodNAME on challenging indoor and outdoor scenarios. Best viewed in color with zoom-in. }
\label{fig:qualitative}
\end{figure*}

\subsection{Kernel Interpolation}
Videos are temporally consistent, which means that adjacent frames usually have strong correlations. This motivates us to apply a key frame propagation strategy. Specifically, 
we only run the whole network on sparse key frames and retrieve the results on other frames by propagating predictions of the key frames. In the context of OV-VIS, we propose a technique called kernel interpolation for fast inference. Suppose we uniformly sample one key frame every $F$ frames ($F=3$ in this work). On these key frames, we run the instance decoder as described in Sec.~\ref{sec:preliminaries} to extract accurate instance kernels for mask prediction. On other non-key frames, instead of running the heavy instance decoder, we interpolate kernels from adjacent key frames to form proxy kernels. Finally, on every frame, the proxy or accurate kernels are convolved with the current pixel embedding map, obtaining the instance masks of the current frame.

For interpolation, we try three different approaches, namely linear, nearest neighbor (NN) and causal nearest neighbor as shown in Fig.~\ref{fig:kernel}. In the linear interpolation, the instance kernels on a non-key frame are the weighted sum of kernels from the adjacent two key frames. Taking frame $T+1$ as an example, the proxy kernels can be computed as $\hat{K}(T+1)=\frac{2}{3}K(T)+\frac{1}{3}K(T+3)$. A simpler way of interpolation is nearest neighbor, which directly copies the kernels from the nearest key frame to the proxy kernels. 

Although these two approaches exploit bi-directional information, it is impossible to access information from the future in real-time applications like autonomous driving. To solve this causality issue, we propose a new interpolation method called causal nearest neighbor. Given a non-key frame, we only consider its causal neighbours, \ie previous key frames. In this way, this approach can not only speed-up the inference but can also be used in online applications. Please note that this technique is training-free, \ie it can still be trained on image-level annotations and does not cause extra complexity. 

To account for the large number of kernels in OV-VIS, we further increase the efficiency of our method by reducing the number of decoder layers from 9 to 3. 

\subsection{Training Strategy}
\label{sec:train-strategy}
Overall, the training process consists of two stages: \textbf{pre-training} on large-scale image datasets; and, \textbf{joint tuning} on many diverse image and video datasets. In the first stage, our model is pre-trained on two large-scale object detection datasets, Objects365~\cite{objects365} and OpenImages~\cite{OpenImages}, which contain 1.8M and 1.7M images respectively. The goal of this stage is to learn the general concept of objects, enabling our model to detect and recognize common instances in the environment. 

Our model in the second stage is fine-tuned on many diverse image and video datasets jointly. Specifically, we additionally introduce image instance segmentation datasets, \eg COCO~\cite{COCO}, LVIS~\cite{LVIS}, VisualGenome~\cite{visualgenome} with rich object-level description and object noun phrases; and, video instance segmentation datasets, \eg YouTube-VIS 2019~\cite{MaskTrackRCNN}, YouTube-VIS 2021~\cite{ytvis21dataset}, and OVIS~\cite{ovis}. 
Consequently, this step gives our model the ability to classify, segment, and track diverse objects in an open environment.

There are some notable differences between our training strategy and that of GLEE~\cite{GLEE}. 
In stage 2, we don't use the datasets without category name~\cite{SAM,UVO}, and the data for referring expression comprehension\&segmentation~\cite{RefCOCOg-umd,RefCOCOandplus,urvos}. 
Moreover, our model is trained for 300K iterations rather than 500K iterations in \cite{GLEE} in both stages.  
Unlike GLEE~\cite{GLEE}, there is no stage 3 of scaling up data~\cite{kosmos2,SAM}, making our method more data-efficient.

\section{Experiments}

\subsection{Implementation Details}
We choose EfficientViT-L2~\cite{EfficientViT} as the vision backbone due to its good balance between accuracy and efficiency. The model is trained on 64 A100 GPUs (40GB) with global batch size of 128 using AdamW~\cite{AdamW} optimizer. The base learning rate is set as $1e^{-4}$. To stabilize training in the beginning, the initial learning is set as $1e^{-8}$ and is then warm-up for 1,000 iterations. Similar to GLEE, the text encoder is frozen in the first stage and fine-tuned in the second stage. During joint tuning, the learning rate of the text encoder is $1/10$ of the base learning rate and an auxiliary distillation loss is introduced to prevent it from drifting too far from the original text embedding space.

\subsection{Qualitative Results}
In this section, we demonstrate some qualitative results of \methodNAME on some challenging scenarios. As shown in Fig.~\ref{fig:qualitative}, \methodNAME can accurately recognize, segment, and track almost all objects in the test scenarios. Even for some uncommon categories like \textit{handle}, \textit{watch}, \textit{pole} and \textit{mitten}, \methodNAME is able to detect and track the instance masks. These results illustrate that \methodNAME can serve as a good open-vocabulary video instance perceiver, which can help robots to better understand dynamic scenes, make decisions and take actions in complex environments.

\subsection{Numerical Evaluation}
To evaluate the open-vocabulary perception ability (including classification, segmentation, and tracking) and the running efficiency of our method, we test it on two large-vocabulary video instance segmentation benchmarks, BURST~\cite{burst} and LV-VIS~\cite{LVVIS}. Following previous works~\cite{GLEE}, we evaluate \methodNAME and compare it with other methods in a zero-shot manner, \ie without learning on the training split of these two datasets.

\paragraph{BURST~\cite{burst}.} 
BURST builds upon the previous large-scale multiple object tracking benchmark TAO~\cite{tao}, extending its original box-level annotations to pixel-precise mask annotations. BURST contains 2,914 videos, including 500, 993, and 1,421 videos in training, validation, and test set, respectively. There are 482 object categories totally, consisting of 425 common categories and 57 uncommon categories. The core ranking metric of BURST~\cite{burst} is Higher Order Tracking Accuracy (HOTA)~\cite{HOTA}, which is the geometric mean of Detection Accuracy (DetA) and the Association Accuracy (AssA). Another metric is mAP, which is the average track-level mask IoU among different categories. 

Based on the measured HOTA, \methodNAME achieves the best results across all, common, and uncommon object categories as shown in Tab.~\ref{tab:video_zeroshot}. Especially on common and all categories, \methodNAME outperforms the previous best method GLEE-Lite by HOTA of 5.9\% and 1.3\%, respectively. Besides, in terms of the segmentation metric mAP, \methodNAME also achieves the best or the second best performance. These results demonstrate \methodNAME's superior video perception ability in challenging scenarios. 

\paragraph{LV-VIS~\cite{LVVIS}.} 
LV-VIS is a more recent benchmark, which includes more videos and covers more diverse object categories. Specifically, the whole dataset includes 4828 videos and there are 3083, 837, and 908 videos in the training, validation, and test set respectively. Besides, LV-VIS covers 1196 object categories, consisting of 641 base categories and 555 novel categories. The final ranking metric is the mean Average Precision (mAP) on all categories, which can also be divided into $\rm AP_b$ on base categories and $\rm AP_n$ on novel categories.

\methodNAME sets the new state-of-the-art performance among efficient OV-VIS methods, surpassing previous best method GLEE-Lite by  1.3\%, 1.3\% and 1.4\% on AP of all, base and novel object categories, respectively. 

\begin{table}[!ht]
\centering
\resizebox{1.0\linewidth}{!}{
\begin{tabular}{lccc}
\toprule
    & \textbf{GLEE-Lite} & \textbf{GLEE-Lite+FEM} & \textbf{\methodNAME} \\
    & LV-VIS (ms) & LV-VIS (ms) & LV-VIS (ms)\\
\midrule
\textbf{Total}   & \textbf{805}  & \textbf{125}    & \textbf{40}              \\
Text Encoder     & 680  &    $<$ 1       & $<$ 1     \\
Feature Enhancer & 75   &   75   & 20              \\
Instance Decoder & 41   &   41   & 4               \\ 
Vision Encoder   & 9    &   9    & 16              \\
\bottomrule
\end{tabular}
}
\caption{Efficiency comparison between the components of GLEE and \methodNAME. GLEE-Lite+FEM means applying Flash Embed Memory (FEM) to GLEE-Lite. The latency is measured on a A100 GPU. The resolution of the input frame is 480p.}
\label{tab:efficiency}
\end{table}


\begin{table*}[!ht]
\centering
\resizebox{0.9\linewidth}{!}{
\begin{tabular}{lcccccccccccc} 
\toprule
\multirow{3}{*}{Method} & \multicolumn{6}{c}{ {\it{BURST~\cite{burst}}}} & \multicolumn{4}{c}{ {\it{LV-VIS~\cite{LVVIS}}}} \\
\cmidrule(lr){2-7} \cmidrule(lr){8-11} 

& \multicolumn{2}{c}{ALL} & \multicolumn{2}{c}{Common} & \multicolumn{2}{c}{Uncommon} & \multirow{2}{*}{$\rm AP$} & \multirow{2}{*}{$\rm AP_b$} & \multirow{2}{*}{$\rm AP_n$} & \multirow{2}{*}{$\rm FPS$} \\
\cmidrule(lr){2-3} \cmidrule(lr){4-5} \cmidrule(lr){6-7}
& $\rm HOTA$ & $\rm mAP$ & $\rm HOTA$ & $\rm mAP$ & $\rm HOTA$ & $\rm mAP$ & & & & \\
\hline
STCN Tracker~\cite{burst} & 5.5 & 0.9 & 17.5 & 0.7 & 2.5 & 0.6 & - & - & - & - \\
Box Tracker~\cite{burst} & 8.2 & 1.4 & 27.0 & 3.0 & 3.6 & 0.9 & - & - & - & - \\
Detic~\cite{detic}-SORT~\cite{bewley2016simple} & - & - & - & - & - & - & 12.8 & 21.1 & 6.6 & 6.7 \\
Detic~\cite{detic}-XMem~\cite{xmem} & - & - & - & - & - & - & 16.3 & 24.1 & 10.6 & 13.4 \\
OV2Seg~\cite{LVVIS} & - & 3.7 & - & - & - & - & 14.2 & 17.2 & 11.9 & 20.1 \\
GLEE-Lite~\cite{GLEE} & 22.6 & \textbf{12.6} & 36.4 & 18.9 & 19.1 & \textbf{11.0} & 19.6 & 22.1 & 17.7 & 1.3 \\
\textbf{\methodNAME} & \textbf{23.9} & 12.4 & \textbf{42.3} & \textbf{19.6} & \textbf{19.3} & 10.7 &\textbf{20.9} &\textbf{23.4} &\textbf{19.1} & \textbf{20.9} \\
\bottomrule
\end{tabular}
}
\caption{Comparison of \methodNAME to recent specialist and generalist models on BURST~\cite{burst} and LV-VIS~\cite{LVVIS} in a zero-shot manner. Evaluation metrics of BURST are reported separately for ‘common’, ‘uncommon’ and ‘all’ classes. The mAP computes mask IoU at the track level, HOTA is a balance of per-frame detection accuracy (DetA) and temporal association accuracy (AssA). The $\rm AP$, $\rm AP_b$, and $\rm AP_n$ in LV-VIS mean the average precision of overall categories, base categories, and novel categories.}
\label{tab:video_zeroshot}
\end{table*}


\begin{table*}[ht]
\centering
\resizebox{1.0\linewidth}{!}{
\begin{tabular}{cccccc|cc}
\toprule
& \textbf{Decoupled Attention} & \textbf{Flash Embed Memory+EVA} & \textbf{Enc3+Dec3} & \textbf{Kernel Interpolation} & \textbf{EfficientViT-L2} & \textbf{Latency (ms)} &  \textbf{AP} \\ \hline
\#1& - & - & - & - & - & 805     & 13.2          \\ 
\#2& \cmark & - & - & - & - & 760  &    13.0          \\ 
\#3& \cmark & \cmark & - & - & - & 80     & 16.3          \\ 
\#4& \cmark & \cmark & \cmark & - & - & 45 & 15.1          \\ 
\#5& \cmark & \cmark & \cmark & \cmark &    - & \textbf{32}        &   14.7          \\ 
\textbf{\#6}& \cmark & \cmark & \cmark & \cmark & \cmark & 40     &   \textbf{15.7}          \\
\bottomrule
\end{tabular}
}
\caption{Ablations for the introduced changes. \cmark means that the corresponding technique is used while dash means not used. The latency and AP performance is tested on LV-VIS benchmark.}
\label{tab:ablation}
\end{table*}


\paragraph{Efficiency.} To demonstrate the efficiency of our method, we compare the overall and component-wise latency of \methodNAME and GLEE-Lite on LV-VIS~\cite{LVVIS} dataset. First, the short side of the input frame is resized to 480 pixels then the resized frame is sent to the network for forward pass. The latency is measured using image batchsize 1 on a A100 GPU. As shown in Tab.~\ref{tab:efficiency}, it takes GLEE-Lite more than 800ms to process one frame and the corresponding text embeddings. However, \methodNAME only needs 40ms, being $\mathbf{20\times}$ faster than GLEE-Lite. We also compare to a variant where we add the Flash Embedding Memory to GLEE-lite, shown as \textbf{GLEE-Lite+FEM} in Fig.~\ref{fig:teaser}. Compared with GLEE-Lite+FEM, \methodNAME still improves the inference speed by $3\times$ thanks to the other introduced methods. 

Each component, the Flash Embedding Memory, decoupled attention, instance decoder and kernel interpolation cumulatively reduce the inference time by a large margin. Notably, \methodNAME spends slightly more compute on the vision encoder where the original ResNet-50~\cite{resnet} backbone is replaced by a stronger EfficientViT-L2~\cite{EfficientViT} backbone. Our ablations in the next section show that although EfficientViT-L2 is slightly slower, the  corresponding performance gain is justifying the compute.




\subsection{Ablation Study}
This section demonstrates the effectiveness of each change to the baseline using the AP on the LV-VIS benchmark as our metric. Since the complete training process described in Sec.~\ref{sec:train-strategy} consumes lots of time and resources, we adopt a more lightweight training setting in the whole ablations. The models in this setting are only trained for 100K iterations on 16 GPUs in both stages rather than 300K iterations on 64 GPUs.

As shown in Tab.~\ref{tab:ablation}, we start with the baseline labeled as \#1, which uses the same architecture as GLEE-Lite but trained with our lightweight training setting. The following \#2 to \#6 incrementally adds new components to highlight their effects on the overall performance. In \#2, we replace the original modality-scale hybrid attention with the more efficient decoupled attention, reducing the latency by 45ms without an obvious performance drop. Then, we introduce Flash Embedding Memory and replace the text encoder with stronger EVA-02-CLIP-L~\cite{eva02} in \#3. This change significantly speeds up our model from 1.3FPS to 12.5FPS and brings an AP improvement of 3.3\%. To further speed up our model, we reduce the number of encoder and decoder layers from \{6,9\} to \{3,3\}. As shown in \#4, although causing a performance drop of 1.2\% AP, this change reduces the latency from 80ms to 45ms (12.5FPS vs 22.2FPS). Furthermore, in \#5, kernel interpolation speeds up our method from 22.2FPS to 31.3FPS, while only causing an AP drop of 0.4\%. Finally, in \#6, by replacing the original vision encoder with stronger EfficientViT-L2 backbone, our final method not only achieves better AP performance but also runs $20\times$ faster than the baseline.

\section{Conclusion}

This work presents \methodNAME, a real-time algorithm for open-vocabulary video instance segmentation, solving the problem of heavy computational costs of current methods. By carefully redesigning the key components of open-vocabulary perception models such as GLEE~\cite{GLEE}, we significantly reduce the processing time per frame while improving or maintaining accuracy across challenging benchmarks. The proposed decoupled attention feature enhancer, Flash Embedding Memory and kernel interpolation techniques are particularly effective in reducing redundant computations and leveraging temporal consistency in video data. Our experiments demonstrate that \methodNAME outperforms existing methods not only in terms of speed but also in accuracy of perceiving objects of diverse categories. These advancements pave the way for broader applications of OV-VIS in fields requiring quick, accurate visual understanding in dynamic, complex scenarios.

\newpage
{\small
\bibliographystyle{ieee_fullname}
\bibliography{egbib}
}

\end{document}